\documentclass[journal]{IEEEtran}

\usepackage{cite}
\usepackage{amsmath,amssymb,amsfonts}
\usepackage{algorithmic}
\usepackage{graphicx}
\usepackage{algorithm,algorithmic}
\usepackage{hyperref}
\hypersetup{hidelinks,pdftitle={Hankel Subspace Self-Supervised Learning for Parallel MRI Reconstruction}}
\usepackage{textcomp}
\usepackage{multirow}
\usepackage{booktabs}
\usepackage{array}
\usepackage{tabularx}
\usepackage{float}
\usepackage{needspace}
\usepackage[table]{xcolor}
\definecolor{tableheader}{RGB}{219,230,239}
\usepackage{stfloats}

\newcommand{\method}{HSSRecon}
\newcommand{\placeholderfigure}[2]{%
    \fbox{%
        \begin{minipage}[c][#1][c]{0.92\columnwidth}
            \centering
            \textcolor{gray}{\textbf{Figure placeholder}}\\[3pt]
            \small\textcolor{gray}{#2}
        \end{minipage}%
    }%
}

\newtheorem{proposition}{Proposition}
\newtheorem{theorem}{Theorem}
\newtheorem{corollary}{Corollary}

\begin{document}

\title{Hankel Subspace Self-Supervised Learning for Parallel MRI Reconstruction}

\author{Mingyu Hu, Siquan Zhu, Xijun Zhong, and Qiegen Liu,
\IEEEmembership{Senior Member, IEEE}
\thanks{This work was supported in part by the National Key Research and
Development Program of China under Grant 2023YFF1204300 and Grant
2023YFF1204302. \textit{(Mingyu Hu and Siquan Zhu contributed equally and share first authorship.)} \textit{(Corresponding author: Qiegen Liu.)}}
\thanks{Mingyu Hu, Xijun Zhong, and Qiegen Liu are with the School of
Information Engineering, Nanchang University, Nanchang, China (e-mail:
\mbox{humingyu0131@gmail.com}, \mbox{416100250216@email.ncu.edu.cn},
\mbox{liuqiegen@ncu.edu.cn}).}
\thanks{Siquan Zhu is with the School of Mathematics and Computer Sciences,
Nanchang University, Nanchang, China (e-mail:
\mbox{zhusiquan@ncu.edu.cn}).}}

\maketitle

\begin{abstract}
Parallel magnetic resonance imaging reconstruction is an ill-posed inverse
problem under undersampling. Multi-coil acquisition and Hankel lifting expose
complementary repeated information: observations of the same anatomy across
coils and repeated local k-space neighborhoods in overlapping windows. These
dependencies guide recovery of missing k-space data. However, splitting lifted
Hankel entries for self-supervision can place the original sample in both input
and target, causing data leakage. We propose Hankel Subspace Self-Supervised
Reconstruction (HSSRecon), a scan-specific reconstruction framework for
parallel magnetic resonance imaging. HSSRecon partitions data by physical
acquisition units before Hankel lifting and applies multiplicity normalization
to repeated Hankel copies in overlapping windows. Rather than learning a
mapping that directly predicts missing data, the network learns a compact
complex-valued Hankel subspace operator. Reconstruction is performed over the
original k-space variables using a conjugate-gradient solver with hard data
consistency. This design separates structural learning in the Hankel domain
from data consistency in the physical domain: the former exploits multi-coil
and local Hankel correlations, while the latter solves over unacquired degrees
of freedom. We provide theoretical analyses of physical-group splitting and
multiplicity normalization, and establish positive definiteness, uniqueness,
hard data consistency, and a finite-step conjugate-gradient error bound for
the system. On fastMRI brain data with three contrasts and three sampling
masks, HSSRecon achieves competitive peak signal-to-noise ratio, structural
similarity, and normalized mean squared error across six aggregated
conditions. Ablations show that physical-group splitting avoids leakage
caused by Hankel-entry splitting, while the learned subspace operator
outperforms a matched fixed operator.
\end{abstract}


\section{Introduction}
Parallel MRI reduces acquisition time by undersampling multi-coil $k$-space.
Reconstruction must recover the missing complex coefficients while preserving
the acquired data and dependencies among coils. The acquisition contains two
complementary forms of redundancy: inter-coil correlations induced by coil
sensitivities and local $k$-space repetitions exposed by overlapping Hankel
lifting. These dependencies are not merely nuisance correlations. They provide
structural constraints that can reduce the ambiguity of the reconstruction.

Classical parallel MRI uses coil information in complementary ways.
SENSE~\cite{pruessmann1999sense} reconstructs with sensitivity maps, whereas
GRAPPA~\cite{griswold2002grappa} interpolates local multi-coil $k$-space
samples. ESPIRiT~\cite{uecker2014espirit} estimates a calibrated signal
subspace, and SPIRiT~\cite{lustig2010spirit} enforces multi-coil
self-consistency. Block-Hankel lifting provides a joint representation of
local multi-coil relations.

Structured low-rank methods use this lift in different ways.
SAKE~\cite{shin2014sake} performs calibrationless matrix completion.
LORAKS~\cite{haldar2014loraks} and P-LORAKS~\cite{haldar2016ploraks}
model local annihilation relations. ALOHA~\cite{jin2016hankel} connects
transform sparsity to low-rank Hankel structure. Other work learns or
streamlines the structural model~\cite{qian2025donate}. Deep-SLR~\cite{pramanik2020deepSLR}
combines structured low-rank iterations with learned regularization.
Convolutional framelet analyses relate learned convolutional representations
to low-rank factorizations~\cite{ye2018deepframelets,zhao2020convframework}.
Separable Hankel regularization reduces the cost of large lifts
\cite{zhang2022sehlr}. SMS-HSL~\cite{park2017smshsl} learns signal and null
subspaces for slice separation, with subsequent work on accelerated T1 mapping
\cite{kim2021smst1,kim2023modelsmshsl}. HKGM~\cite{peng2023hkgm} learns a
one-shot generative prior in Hankel $k$-space for parallel imaging. These
studies establish that inter-coil and local Hankel redundancy can provide
useful reconstruction priors. Most of them, however, use calibration, explicit
matrix completion, or iterative prior optimization rather than learning the
structural constraint from the same scan through a self-supervised target.

Model-based networks show how a learned regularizer can be coupled to the
acquisition model~\cite{hammernik2018variational,aggarwal2019modl}, while
RAKI~\cite{akcakaya2019raki} learns scan-specific interpolation from acquired
calibration data. Self-supervised reconstruction instead forms training
targets from the acquired measurements themselves~\cite{joo2025aespa}. SSDU~\cite{yaman2020ssdu}
separates input and loss samples. Multi-mask SSDU~\cite{yaman2022multimask}
repeats that split to obtain more training views. Zero-shot
self-supervision~\cite{yaman2021zsssl} adapts to one scan with reference-free
selection. Deep unfolding informed by matrix completion has also been used for
self-supervised $k$-space interpolation~\cite{luo2025hsspgd}. These methods
define their input and target roles in the original acquisition coordinates,
whereas structured Hankel modeling provides a complementary prior for
plausible $k$-space reconstructions. Their combination is nontrivial because
overlapping Hankel lifting duplicates physical coordinates.

\begin{figure}[!h]
    \centering
    \includegraphics[width=\columnwidth]{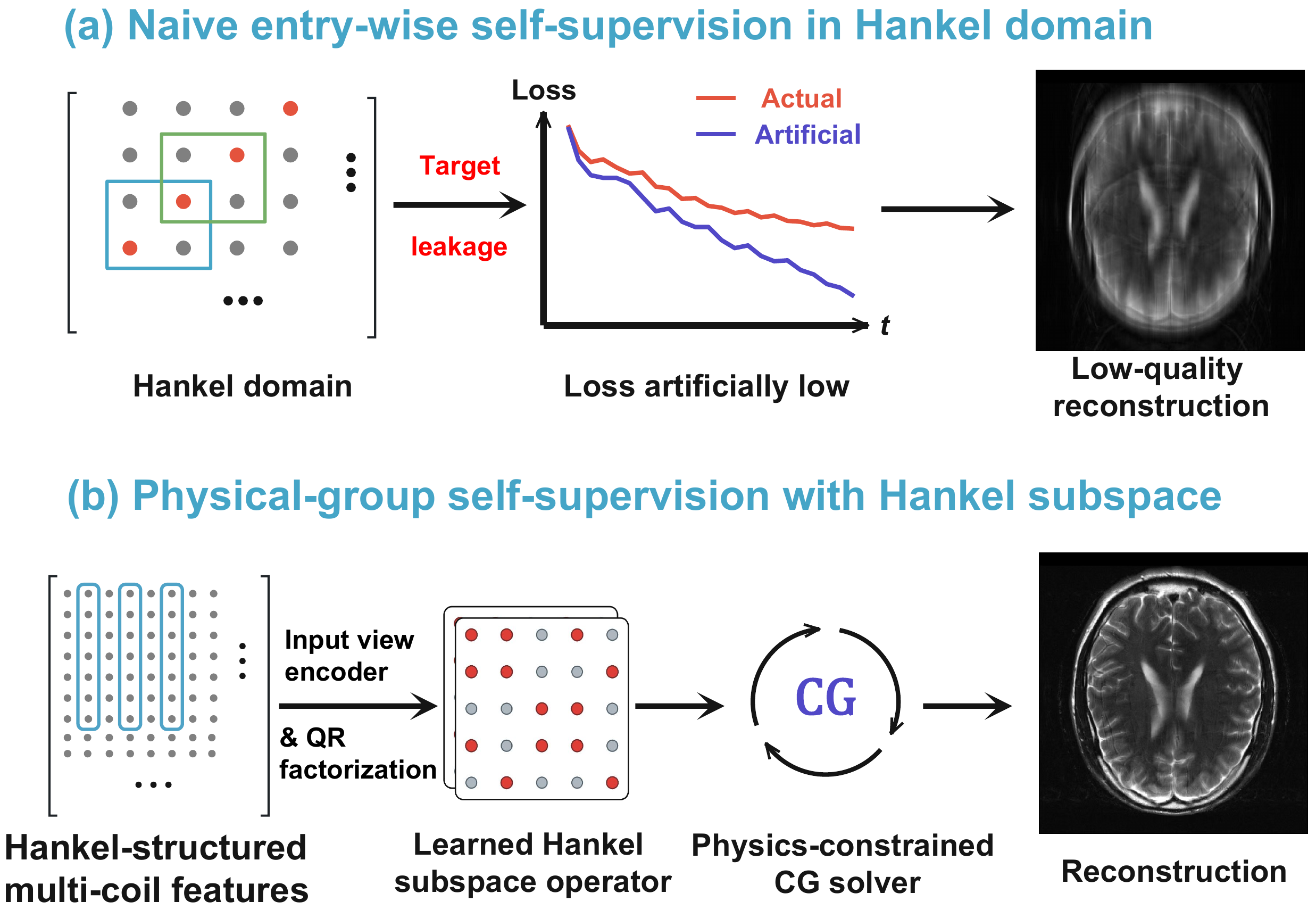}%
    \caption{Hankel-domain leakage and leakage-safe physical-group
    self-supervision. (a) Naive entry-wise splitting after lifting can expose
    held-out coefficients through duplicated entries, producing an artificially
    low loss. (b) Physical-group splitting enables a learned Hankel subspace
    operator and a physics-constrained CG solver for reconstruction.}
    \label{fig:comparison}
\end{figure}

A key requirement of self-supervised reconstruction is to ensure that
measurements used as targets cannot be accessed by the predictor. Earlier
self-supervised formulations address this issue by explicitly defining which
acquired samples are used as inputs and which are reserved as targets
~\cite{batson2019noise2self,aggarwal2023ensure,millard2022noisier2noise,
wang2024kband,xu2026units}. In the present setting,
Hankel lifting does not preserve a one-to-one correspondence between physical
coefficients and lifted entries: one physical coefficient may occupy several
entries after overlapping windows are formed. Splitting entries after lifting
can therefore leave a copy of a held-out target in the input. Moreover, a
lifted loss without multiplicity correction gives greater weight to
coefficients that appear more often. A valid self-supervised formulation with
an overlapping Hankel lift must therefore both exclude targets at the
level of physical acquisition and account for repeated coordinates, as illustrated
in Fig.~\ref{fig:comparison}(a).

The central design question is how to retain inter-coil and local Hankel
redundancy without allowing repeated representations to leak targets or alter
the physical loss. Hankel Subspace Self-Supervised Reconstruction (HSSRecon)
addresses this question by assigning input, held-out, and validation roles to
physical acquisition units before overlapping windows are formed and correcting
repeated Hankel copies with multiplicity normalization. It then learns a
Hankel subspace operator and solves for the physical $k$-space variables with a
physics-constrained conjugate-gradient (CG) solver.
Fig.~\ref{fig:comparison} summarizes this contrast: entry-wise splitting after
lifting can leak held-out coefficients in (a), whereas physical-group
self-supervision learns a Hankel subspace operator and uses a
physics-constrained CG solver in (b).

The main contributions are summarized as follows:
\begin{itemize}
    \item \textbf{\textit{We introduce HSSRecon, a Hankel-subspace
    self-supervised reconstruction framework for parallel MRI.}} It learns a
    compact subspace from multi-coil $k$-space redundancy and couples it with
    a CG solver with hard data consistency.
    \item \textbf{\textit{We develop a leakage-safe physical-group
    supervision scheme for overlapping Hankel lifts.}} Splitting before
    lifting excludes duplicated copies of held-out coefficients, while
    multiplicity normalization preserves the physical loss.
    \item \textbf{\textit{We provide theoretical and empirical validation
    of HSSRecon.}} The analysis establishes well-posed reconstruction and
    finite-step CG bounds, while experiments demonstrate competitive quality
    across the evaluated multi-coil datasets and sampling masks.
\end{itemize}

\section{Method}

\subsection{Overview}
Let $x\in\mathbb{C}^{H\times W}$ denote the object and let
$s_c\in\mathbb{C}^{H\times W}$ be the sensitivity map of coil $c$.
The acquired multi-coil $k$-space is:
\begin{equation}
    y_c = M_{\Omega}\odot\mathcal{F}\{s_c x\}+\varepsilon_c,
    \qquad c=1,\ldots,C,
    \label{eq:forward}
\end{equation}
where $M_{\Omega}$ is the physical acquisition mask, $\Omega$ is its support,
$\mathcal{F}$ is the centered orthonormal Fourier transform, and
$\varepsilon_c$ denotes measurement noise supported on $\Omega$. We write
$y=(y_1,\ldots,y_C)$ for the multi-coil measurement. For any support $S$,
$P_S$ is the coordinate projector that retains entries in $S$ and sets all
other entries to zero, so $P_S y$ has the same dimensions as $y$.

\begin{figure*}[!t]
    \centering
    \includegraphics[width=\textwidth]{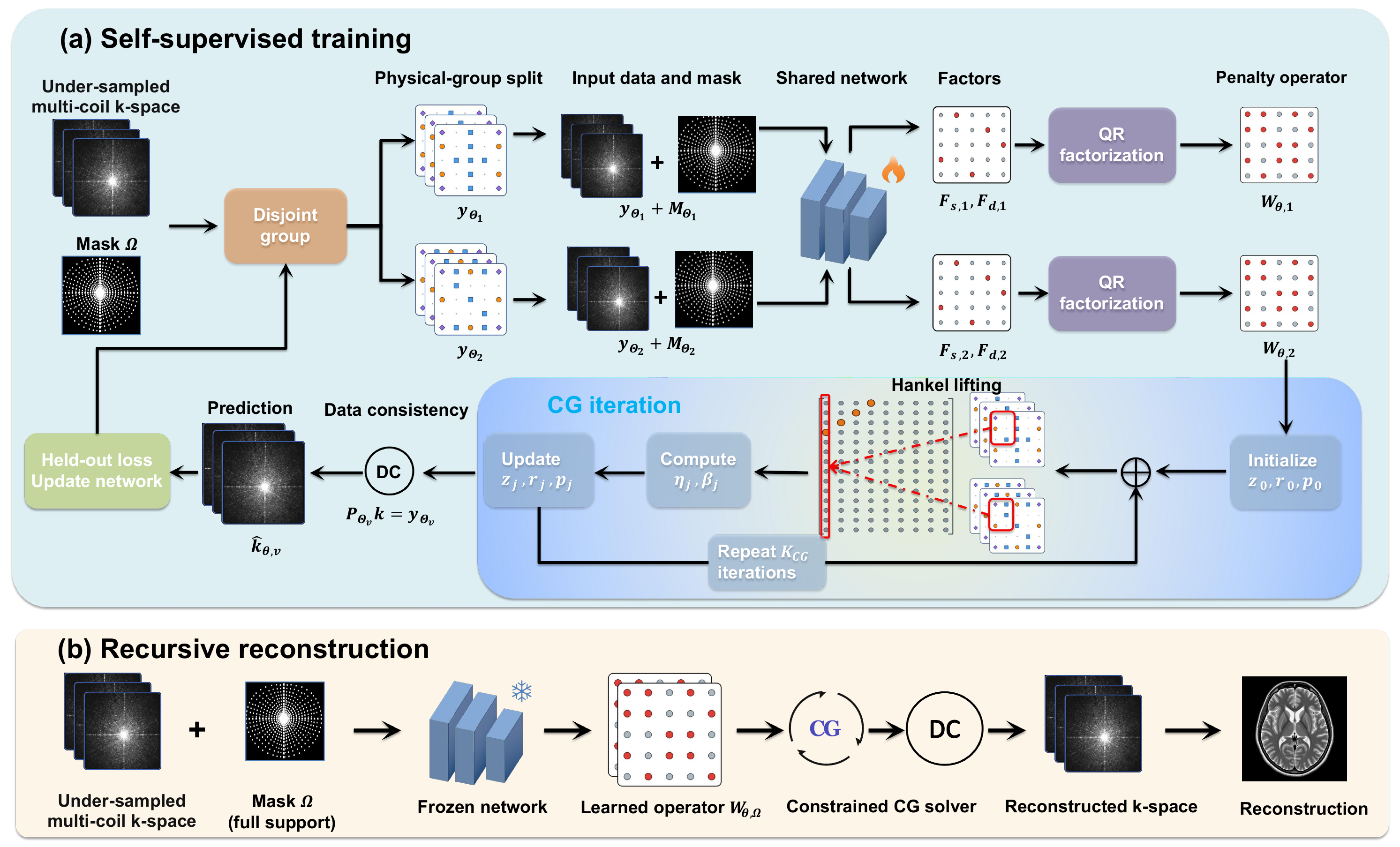}%
    \caption{Overview of HSSRecon. (a) During self-supervised training, the
    acquired multi-coil $k$-space and mask are split into disjoint physical
    groups, processed by the shared learner to produce factorized subspaces, and
    used in a constrained Hankel CG loop. The learner is updated using the
    physical held-out loss. (b) During reconstruction, the trained learner is
    frozen. It uses the full acquired data and mask to construct the subspace
    operator, after which the constrained CG solver reconstructs the missing
    $k$-space coefficients,
    followed by inverse Fourier imaging.}
    \label{fig:pipeline}
\end{figure*}

The HSSRecon framework has three functional components. First, leakage-safe
physical-group self-supervision defines the input, held-out, and validation
views before Hankel lifting. Second, the Hankel subspace learner maps visible
multi-coil data and mask geometry to a compact complex operator. Third, the
physics-constrained reconstruction solver uses this operator to recover
missing coefficients while retaining all acquired samples. We update the shared
learner using the held-out prediction
loss and select the checkpoint using the validation groups. At deployment, the
selected learner operates on the full acquired support.
Fig.~\ref{fig:pipeline} and Algorithm~\ref{alg:hssrecon} summarize the
workflow.

\subsection{Hankel Subspace Self-Supervision}
\begin{figure}[!h]
    \centering
    \includegraphics[width=0.99\columnwidth]{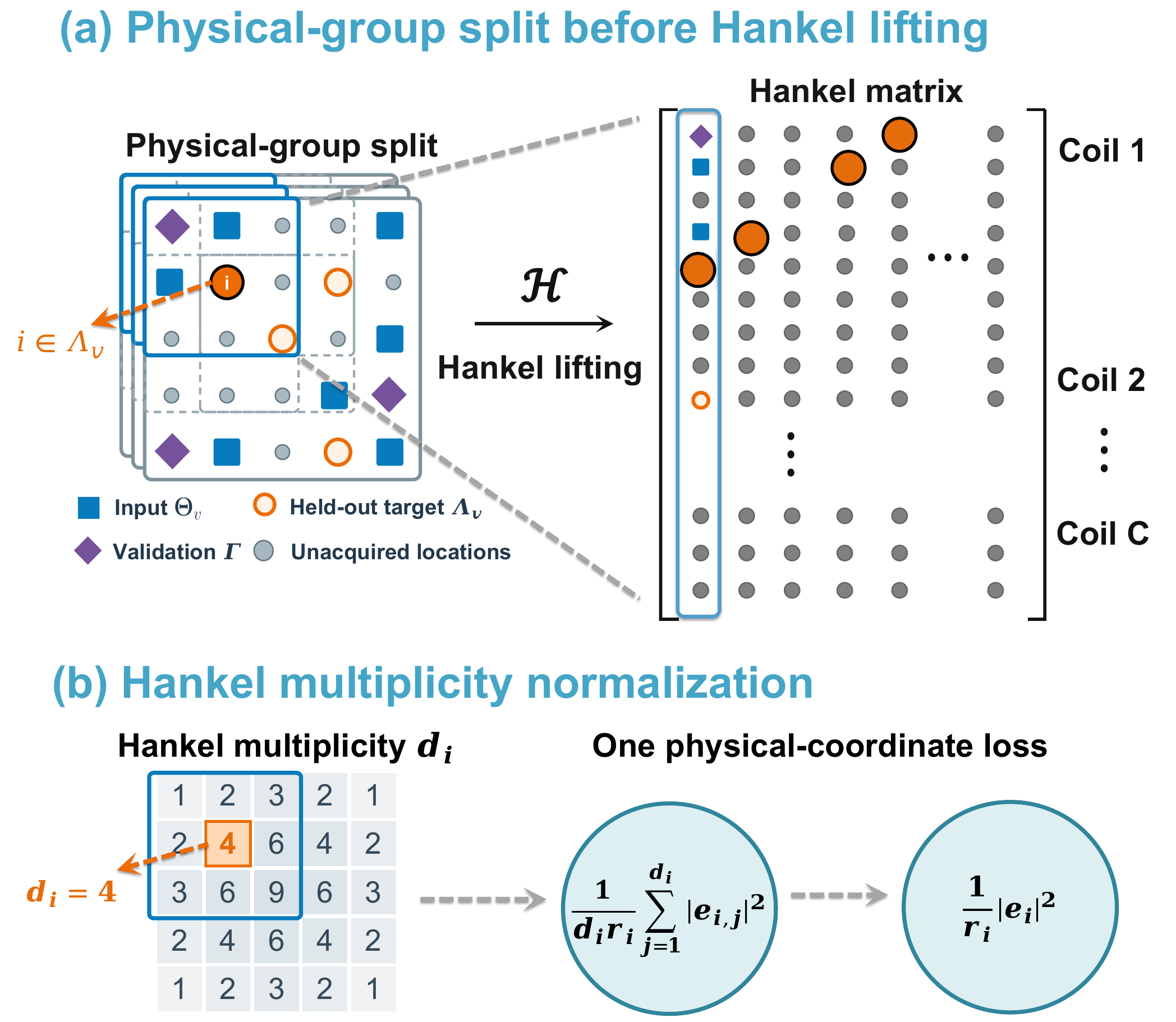}%
    \caption{Leakage-safe physical grouping and Hankel multiplicity
    normalization. (a) Acquired multi-coil data are split into input
    $\Theta_v$, held-out $\Lambda_v$, and validation $\Gamma$ groups before
    Hankel lifting. (b) After projection onto $\operatorname{range}(\mathcal H)$,
    the $d_i$ copies of $k_i$ satisfy
    $\frac{1}{r_i}\left(\frac{1}{d_i}\sum_j|e_{i,j}|^2\right)
    =\frac{1}{r_i}|e_i|^2$.
    The global factors $1/V$ and $1/Z_v$ are omitted for clarity.}
    \label{fig:physical-split}
\end{figure}

Self-supervision is defined on physical acquisition units rather than on
individual entries of a lifted matrix. An acquired phase-encoding line, a
spoke-shaped set of Cartesian-grid samples, or a sampled spatial coordinate
shared across coils is treated as one unit. This choice retains inter-coil
measurements as a single observation and prevents a held-out coefficient from
reappearing through overlapping windows. The resulting supervision uses two
operations: physical-group splitting before Hankel lifting and multiplicity
normalization of the repeated Hankel copies, as summarized in
Fig.~\ref{fig:physical-split}.

Let $\mathcal{G}=\{g_1,\ldots,g_G\}$ partition the acquired support into
physical sampling units. We use acquired phase-encoding lines for uniform
masks, disjoint spoke-shaped Cartesian-grid groups for radial masks, and
individual acquired spatial coordinates for Poisson masks. Each group contains
the corresponding measurements from all coils. For view $v$, the groups are
split into:
\begin{equation}
    \mathcal{G}
    =
    \mathcal{G}_{\Theta_v}\mathbin{\dot\cup}
    \mathcal{G}_{\Lambda_v}\mathbin{\dot\cup}
    \mathcal{G}_{\Gamma},
    \qquad
    \Theta_v\cap\Lambda_v=\varnothing,
    \label{eq:physical_split}
\end{equation}
where $\Theta_v$, $\Lambda_v$, and $\Gamma$ are the unions of input, held-out,
and validation groups, respectively. We draw $\Gamma$ once before training.
Only the remaining groups are repartitioned into $\Theta_v$ and $\Lambda_v$
at each update. Splitting before lifting excludes
every Hankel copy of a held-out coefficient from the input. The Hankel
subspace learner receives
$y_{\Theta_v}=P_{\Theta_v}y$ and mask descriptors. $\Lambda_v$ supplies
training targets, and $\Gamma$ is reserved for reference-free model selection.

\subsubsection{Hankel Multiplicity Normalization}
Let $\mathcal{H}$ be a block-Hankel lifting operator that extracts overlapping
$k_h\times k_w$ neighborhoods from all coils:
\begin{equation}
    \mathcal{H}(k)
    =
    \begin{bmatrix}
    \mathcal{E}_1(k)^{\mathsf T}\\
    \vdots\\
    \mathcal{E}_N(k)^{\mathsf T}
    \end{bmatrix}
    \in\mathbb{C}^{N\times F},
    \qquad
    F=Ck_hk_w.
    \label{eq:hankel}
\end{equation}
Each coefficient $k_i$ at one coil and one $k$-space location appears in $d_i$
overlapping windows. For this lifting construction,
$D=\mathcal{H}^{*}\mathcal{H}$ is diagonal with
$D_{ii}=d_i>0$. Define:
\begin{equation}
    \overline{\mathcal{H}}=\mathcal{H}D^{-1/2},
    \qquad
    \mathcal{H}^{\dagger}=D^{-1}\mathcal{H}^{*},
    \qquad
    P_{\mathcal{H}}=\mathcal{H}D^{-1}\mathcal{H}^{*}.
    \label{eq:quotient}
\end{equation}
Here $\mathcal{H}^{\dagger}Z$ averages overlapping copies of a lifted estimate
$Z$ back to physical coordinates, and $P_{\mathcal{H}}$ projects it onto the range of
the lift. For a held-out support $\Lambda$:
\begin{equation}
    \sum_{a\in\mathcal{C}(\Lambda)}
    \frac{\left|[P_{\mathcal{H}}Z]_a-[\mathcal{H}y]_a\right|^2}
    {d_{\pi(a)}\,r_{\pi(a)}}
    =
    \sum_{i\in\Lambda}
    \frac{\left|[\mathcal{H}^{\dagger}P_{\mathcal{H}}Z]_i-y_i\right|^2}
    {r_i},
    \label{eq:quotient_identity}
\end{equation}
Here $\mathcal{C}(\Lambda)$ contains all lifted copies of coordinates in
$\Lambda$, $\pi(a)$ identifies the physical coordinate of copy $a$, and
$r_i>0$ is its pre-draw probability of target inclusion. For the $j$th lifted
copy $a_{i,j}$ of coordinate $i$, define
$e_{i,j}=[P_{\mathcal H}Z]_{a_{i,j}}-[\mathcal{H}y]_{a_{i,j}}$ and
$e_i=[\mathcal{H}^{\dagger}P_{\mathcal H}Z]_i-y_i$. Because
$P_{\mathcal H}Z$ lies in the range of $\mathcal H$, the copies have the same
residual after projection, $e_{i,1}=\cdots=e_{i,d_i}=e_i$. Thus, the factor
$1/d_i$ removes repeated counting, while $1/r_i$ retains the pre-draw
weighting by target inclusion. For a uniform fixed-size draw of $K$ target
groups from $\mathcal{G}_{\mathrm{train}}$, which contains the groups not
reserved for validation,
$r_i=K/|\mathcal{G}_{\mathrm{train}}|$.

For a Hankel-consistent lifted estimate, obtained after projection onto the
range of $\mathcal{H}$, the $d_i$ copies share the same residual. Thus, their
weighted squared errors reduce to one physical-coordinate contribution, as
illustrated in Fig.~\ref{fig:physical-split}(b). The same correction applies to
the two-dimensional multi-coil lift used here.

\Needspace{13\baselineskip}
\subsubsection{Physical Held-Out Objective}
For view $v$, let $\widehat{k}_{\theta,v}$ denote the reconstruction produced
from $y_{\Theta_v}$ and $W_{\theta,v}$. The self-supervised loss is evaluated
on the physical held-out coordinates:
\begin{equation}
    \mathcal{L}_{\mathrm{SSL}}(\theta)
    =
    \frac{1}{V}\sum_{v=1}^{V}
    \frac{1}{Z_v}
    \sum_{i\in\Lambda_v}
    \frac{1}{r_i}
    \left\|\widehat{k}_{\theta,v,i}-y_i\right\|_2^2,
    \label{eq:ssl_loss}
\end{equation}
where $r_i$ is fixed by the sampling design before the view is drawn, and
$Z_v>0$ sets the optimization scale. In the reported two-view implementation,
the two weighted sums are divided by the common factor
$2C|\Lambda_1|$. Thus $V=2$ and $Z_v=C|\Lambda_1|$ for both views. The loss is
evaluated in physical $k$-space. Eq.~\eqref{eq:quotient_identity} gives its
equivalent multiplicity-corrected lifted form. Held-out values enter only
the loss, while the Hankel prior enters the constrained solve below. The
normalization sets the optimization scale.
For the reported two-view training, we also penalize disagreement between
the two predicted operators:
\begin{equation}
    \mathcal{L}_{\mathrm{train}}
    =\mathcal{L}_{\mathrm{SSL}}
    +\lambda_{\mathrm{cv}}
    \frac{\|W_{\theta,1}-W_{\theta,2}\|_F^2}{F^2}.
    \label{eq:training_loss}
\end{equation}
This term compares the two operators without using held-out values.

\subsection{Hankel Subspace Learner}
The Hankel subspace learner maps the real and imaginary parts of
$y_{\Theta_v}$, the input mask, and a descriptor of the geometry of the full
acquired mask to a scan-specific embedding. The descriptor contains sampling
locations but no held-out values. A convolutional trunk with a $5\times5$
input layer and $3\times3$ blocks consisting of convolution, normalization,
and GELU operations, followed by global pooling, produces a
scan-specific embedding. A second encoder adds mask statistics. Complex
heads output the structural ($s$) and sampling-conditioned innovation ($d$)
factors:
\begin{equation}
    F_{s,v}\in\mathbb{C}^{F\times q_s},
    \qquad
    F_{d,v}\in\mathbb{C}^{F\times q_d},
    \label{eq:factors}
\end{equation}
where $q_s$ and $q_d$ are their respective column counts. The reported
configuration uses $q_s=16$ and $q_d=8$.

For full-column-rank factors, $F_{s,v}R$ spans the same subspace for any
invertible $R$. A reduced complex QR decomposition then gives the structural
projector:
\begin{equation}
    F_{s,v}=Q_{s,v}R_{s,v},
    \qquad
    P_{s,v}=Q_{s,v}Q_{s,v}^{H}.
    \label{eq:projector}
\end{equation}
We remove the structural component from the innovation factor before its QR
decomposition:
\begin{equation}
    \begin{aligned}
        \widetilde{F}_{d,v}
        &= (I-P_{s,v})F_{d,v},\\
        \widetilde{F}_{d,v}
        &= Q_{d,v}R_{d,v},\qquad
        P_{d,v}=Q_{d,v}Q_{d,v}^{H}.
    \end{aligned}
    \label{eq:innovation_projector}
\end{equation}
We combine the structural and innovation projectors to define the operator:
\begin{equation}
    W_{\theta,v}=P_{s,v}+\alpha_v P_{d,v},
    \qquad
    0\leq \alpha_v\leq \alpha_{\max}.
    \label{eq:operator}
\end{equation}
We set $\alpha_{\max}=0.1$ and compute $\alpha_v$ by multiplying it by the
two sigmoid outputs for innovation weighting and confidence gating. The
resulting $W_{\theta,v}$ is Hermitian positive
semidefinite. When both factors entering QR have full column rank, it depends
on their subspaces rather than their arbitrary factor coordinates. This
invariance is not asserted for rank-deficient factors.

The network processes the real and imaginary input channels to produce complex
factors, and
$(\cdot)^H$ denotes Hermitian transpose. The feature dimension
$F=Ck_hk_w$ counts the entries in a $k_h\times k_w$ neighborhood across $C$
coils. In the quadratic
below, $W_{\theta,v}$ weights filter directions with residual Hankel energy.
It is not a projector onto a reconstructed signal subspace.

\subsection{Physics-Constrained Reconstruction Solver}
For a fixed $W_{\theta,v}$, we formulate reconstruction as the following
constrained quadratic problem over multi-coil $k$-space:
\begin{equation}
    \begin{aligned}
        \widehat{k}_{\theta,v}^{(\infty)}
        =\arg\min_{k}\;&
        \frac{\lambda_H}{2}
        \left\|\overline{\mathcal{H}}(k)W_{\theta,v}^{1/2}\right\|_F^2\\
        &+\frac{\lambda_0}{2}\left\|k\right\|_2^2
        \quad\text{s.t.}\quad
        P_{\Theta_v}k=P_{\Theta_v}y .
    \end{aligned}
\label{eq:inner_problem}
\end{equation}
Here $\lambda_H\geq0$ weights the learned Hankel penalty and $\lambda_0>0$
is a ridge term. The first term penalizes energy along the filter directions
selected by $W_{\theta,v}$. The constraint retains all input measurements.

The normal map uses $\overline{\mathcal{H}}=\mathcal{H}D^{-1/2}$, so a
coefficient appearing in more windows does not gain weight solely from its
copy count. We absorb the normal map's fixed scalar normalization into
$\lambda_H$.

Let $N_v$ contain an orthonormal basis of the free-coordinate space
$\ker P_{\Theta_v}$, and set $k_{\mathrm{dc},v}=P_{\Theta_v}y$. Every
feasible reconstruction has the form $k=k_{\mathrm{dc},v}+N_vz$, where $z$
contains only unmeasured coefficients. Define the feature-space action
$\mathcal{W}_{\theta,v}(Z)=ZW_{\theta,v}$ and the physical normal operator
$R_{\theta,v}$ and right-hand side $c_v$ by:
\begin{equation}
    R_{\theta,v}=\lambda_H\overline{\mathcal{H}}^{*}
    \mathcal{W}_{\theta,v}\overline{\mathcal{H}}+\lambda_0 I,
    \qquad
    c_v=-N_v^{H}R_{\theta,v}k_{\mathrm{dc},v}.
    \label{eq:hessian_and_rhs}
\end{equation}
The reduced normal matrix is:
\begin{equation}
    A_v=N_v^{H}
    R_{\theta,v}N_v
    \succ 0.
\label{eq:normal}
\end{equation}
CG solves $A_vz=c_v$ in the free-coordinate space. With
$r_0=c_v-A_vz_0$ and $p_0=r_0$, its iterations are:
\begin{equation}
    \begin{aligned}
        r_j&=c_v-A_vz_j,\qquad
        z_{j+1}=z_j+\eta_jp_j,\\
        p_{j+1}&=r_{j+1}+\beta_jp_j .
    \end{aligned}
    \label{eq:cg_update}
\end{equation}
where $\eta_j$ and $\beta_j$ are the standard CG scalars. After at most
$K_{\mathrm{CG}}$ iterations, it returns
$\widehat{k}_{\theta,v}^{(K_{\mathrm{CG}})}
 =k_{\mathrm{dc},v}+N_vz_{K_{\mathrm{CG}}}$. This parametrization preserves
hard data consistency at every iteration. The implementation also writes the
acquired values back to the returned array to suppress numerical drift.

\begin{algorithm}[!t]
\caption{HSSRecon}
\label{alg:hssrecon}
\begin{algorithmic}[1]
\REQUIRE Acquired multi-coil $k$-space $y$, support $\Omega$, physical groups
$\mathcal{G}$, training updates $T$, and CG budget $K_{\mathrm{CG}}$
\ENSURE Reconstructed multi-coil $k$-space $\widehat{k}$ and image
\STATE Fix validation groups $\mathcal{G}_{\Gamma}$ and initialize $\theta$
\FOR{$t=1,\ldots,T$}
    \STATE For $v=1,2$, partition $\mathcal{G}\setminus\mathcal{G}_{\Gamma}$
    into $\mathcal{G}_{\Theta_v}$ and $\mathcal{G}_{\Lambda_v}$ before lifting
    \FOR{$v=1,2$}
        \STATE Form $y_{\Theta_v}=P_{\Theta_v}y$ and predict
        $W_{\theta,v}$ from visible values and mask geometry
        \STATE Initialize $z_0=0$ and run up to $K_{\mathrm{CG}}$ steps on
        $A_vz=c_v$ and set $\widehat{k}_{\theta,v}
        =k_{\mathrm{dc},v}+N_vz_{K_{\mathrm{CG}}}$
        \STATE Evaluate the physical held-out loss on $\Lambda_v$
    \ENDFOR
    \STATE Add cross-view operator consistency, backpropagate
    $\mathcal{L}_{\mathrm{train}}$, and update $\theta$
    \STATE Select checkpoints using only validation support $\Gamma$
\ENDFOR
\STATE Freeze the selected learner and selection rule
\STATE \textbf{Iterative reconstruction:} predict $W_{\theta,\Omega}$ from the full acquired
support and run constrained CG
\STATE Restore acquired values and return $\widehat{k}$ and its
inverse Fourier image
\end{algorithmic}
\end{algorithm}

\subsection{Theoretical Analysis}
We establish the effect of physical splitting and multiplicity correction,
then analyze the constrained reconstruction for a fixed view and operator.

\begin{samepage}
\begin{proposition}[\textbf{Physical splitting and multiplicity correction}]
\label{prop:quotient}
For a pure lift that copies each physical entry and has $d_i>0$, the physical
split in
Eq.~\eqref{eq:physical_split} excludes every lifted copy of a held-out value
from the current view's input. For any lifted estimate $Z$,
Eq.~\eqref{eq:quotient_identity} counts each physical-coordinate error once
after projection onto the lift's range and multiplicity correction.
The normalized lift also satisfies:
\begin{equation}
    \overline{\mathcal H}^{*}\overline{\mathcal H}=I,
    \qquad \|\overline{\mathcal H}k\|_F^2=\|k\|_2^2.
    \label{eq:lift_isometry}
\end{equation}
\end{proposition}
\end{samepage}

This isometry links multiplicity correction to the solver's spectral scale.
For the orthogonal structural and innovation subspaces in
Eqs.~\eqref{eq:projector}--\eqref{eq:operator}, assuming full column rank at
both QR steps, the eigenvalues of $W_{\theta,v}$ belong to
$\{0,1,\alpha_v\}$. Hence its spectral norm
$w_v=\|W_{\theta,v}\|_2$ is bounded by $\max\{1,\alpha_{\max}\}$.
This bound limits the penalty strength without requiring the learned filter
directions to match the signal.

\begin{theorem}[\textbf{Constrained reconstruction}]
\label{thm:solver}
Fix view $v$ and $W_{\theta,v}\succeq0$, with $\lambda_H\geq0$ and
$\lambda_0>0$. Then $A_v$ is Hermitian positive definite,
Eq.~\eqref{eq:inner_problem} has a unique minimizer, and every iterate
$k^{(j)}=k_{\mathrm{dc},v}+N_vz_j$ satisfies hard data consistency.
With $w_v=\|W_{\theta,v}\|_2$, the normalized lift gives:
\begin{equation}
    \begin{aligned}
    \lambda_0 I&\preceq A_v\preceq(\lambda_0+\lambda_Hw_v)I,\\
    \kappa_v&\leq 1+\frac{\lambda_Hw_v}{\lambda_0}.
    \end{aligned}
    \label{eq:solver_spectrum}
\end{equation}
Here $\mu_v=\lambda_{\min}(A_v)$,
$\kappa_v=\lambda_{\max}(A_v)/\mu_v$, and $e_0=z_0-A_v^{-1}c_v$.
With $\|u\|_{A_v}=(u^HA_vu)^{1/2}$ and
$\delta_{\mathrm{CG}}(K)=\|\widehat{k}_{\theta,v}^{(K)}-
\widehat{k}_{\theta,v}^{(\infty)}\|_2$, exact-arithmetic CG satisfies:
\begin{equation}
    \delta_{\mathrm{CG}}(K)
    \leq\frac{2}{\sqrt{\mu_v}}
    \left(\frac{\sqrt{\kappa_v}-1}{\sqrt{\kappa_v}+1}\right)^K
    \|e_0\|_{A_v}.
    \label{eq:cg_bound}
\end{equation}
For the final residual $r_K=c_v-A_vz_K$, the remaining solver error also
admits the computable bound:
\begin{equation}
    \delta_{\mathrm{CG}}(K)
    \leq\frac{\|r_K\|_2}{\mu_v}
    \leq\frac{\|r_K\|_2}{\lambda_0}.
    \label{eq:cg_residual_bound}
\end{equation}
\end{theorem}

\begin{corollary}[\textbf{Reconstruction error bound}]
\label{cor:reconstruction-error}
Under the conditions of Theorem~\ref{thm:solver}, for any evaluation
reference $k_\star$, define the feasible comparison point and its Hankel
penalty response on free coordinates:
\[
\begin{aligned}
k_{\mathrm{ref}}&=P_{\Theta_v}y+(I-P_{\Theta_v})k_\star,\\
g_{H,v}(k_{\mathrm{ref}})&=N_v^H\overline{\mathcal H}^{*}
\mathcal W_{\theta,v}(\overline{\mathcal H}k_{\mathrm{ref}}).
\end{aligned}
\]
The finite-step reconstruction obeys:
\begin{equation}
    \begin{aligned}
    \|\widehat{k}_{\theta,v}^{(K)}-k_\star\|_2
    \leq{}&\|P_{\Theta_v}(y-k_\star)\|_2\\
    &+\frac{\lambda_H}{\mu_v}\|g_{H,v}(k_{\mathrm{ref}})\|_2\\
    &+\frac{\lambda_0}{\mu_v}\|N_v^H k_{\mathrm{ref}}\|_2
      +\delta_{\mathrm{CG}}(K).
    \end{aligned}
    \label{eq:total_bound}
\end{equation}
\end{corollary}

The spectral bound in Eq.~\eqref{eq:solver_spectrum} has no explicit
 multiplicity factor. The corresponding unnormalized bound contains
$\max_i d_i$. The exact-arithmetic contraction in Eq.~\eqref{eq:cg_bound} and
the residual bound in Eq.~\eqref{eq:cg_residual_bound} quantify error relative
to the exact solution of the fixed quadratic.

The four terms in Eq.~\eqref{eq:total_bound} account for input measurement
discrepancy, Hankel and ridge penalty responses at $k_{\mathrm{ref}}$, and
incomplete solution. Any discrepancy between the acquired measurements and
$k_\star$ remains in every data-consistent reconstruction.
The ridge term can also bias free coefficients even when the Hankel response
vanishes. These bounds apply to a fixed view and operator. They do not
establish recovery of the true signal subspace or convergence of network
training. Proofs are given in the Appendix.

\section{Experiments}
\label{sec:experiments}

\subsection{Experimental Setup}
\label{subsec:experimental-setup}
\textbf{\textit{Datasets:}}
We evaluate the proposed method on two fully sampled multi-coil benchmarks:
fastMRI brain~\cite{knoll2020fastmri,muckley2021fastmriChallenge} and an
independent 10-subject panel selected from the public Stanford 2D FSE dataset
~\cite{cheng2018stanford2dfse,fabian2021mraugment}. Both datasets provide complex
multi-coil $k$-space. For each case, all methods use the same multi-coil
measurements and undersampling mask. Fully sampled data are used only to form
the reference RSS images, compute quantitative scores, and produce visualizations
after training and model selection.
The fastMRI panel contains three contrasts (AXFLAIR, AXT1PRE, and AXT2), five
subjects per contrast, and slices 003, 008, and 012. Its working grid is
$640\times320$ with 16 or 20 receiver coils. We retrospectively apply
Poisson, radial, and uniform masks at $R\in\{6,8\}$, yielding
$3\times5\times3\times3\times2=270$ cases.

The Stanford panel contains 10 held-out subjects and three interior slices per
subject. Unlike fastMRI, it represents a single 2D FSE Bone acquisition
setting rather than three contrasts, with 16 receiver coils. It uses the same
three mask families at $R=4$, giving 30 cases per mask and 90 cases in total.
The data retain native full-FOV matrices rather than a common working grid:
the first spatial dimension ranges from 174 to 256, and the second is either
320 or 352. This variable native geometry provides a complementary test of
transfer across acquisition and anatomy.

\textbf{\textit{Compared methods:}}
We compare zero filling (ZF), calibrationless SAKE~\cite{shin2014sake},
ALOHA~\cite{jin2016hankel}, SSDU~\cite{yaman2020ssdu}, BOINR
~\cite{yu2026bilevelinr}, and \method. ZF, SAKE, and ALOHA do not use
external training data. SSDU is a reference-free self-supervised method
trained at the dataset level. One model is trained for each contrast, mask,
and acceleration factor in the fastMRI experiment. Each model uses five
training subjects that do not overlap with the test subjects and 50 training
slices. In the Stanford experiment, one model is trained for each mask using
five such training subjects and 55 training slices. BOINR and \method~optimize
their parameters independently for each test scan using its acquired
measurements.
The reported BOINR results use RSS-based phase estimation from acquired ACS
data, followed by acquired-data hard consistency. These choices are fixed
before scoring. We
verify physical support, duplicate handling, hard data consistency, and solver
residuals for each case.

\textbf{\textit{Implementation details:}}
\method~uses a $3\times3$ Hankel lift, factorized structural and innovation
subspaces, two physical training views, a 24-step CG budget,
$\lambda_H=1$, $\lambda_0=10^{-3}$, $\lambda_{\mathrm{cv}}=0.1$, and hard
consistency on acquired samples. The fastMRI comparison uses 300 updates for
Poisson and radial masks and 1000 updates for the uniform mask. The Stanford
panel uses prespecified iteration budgets of 300, 600, and 1000 updates for
Poisson, radial, and uniform masks, respectively. Checkpoint selection withholds
the $\Gamma$ groups and uses the larger loss from two partitions formed with the
same physical-group rule used during reconstruction. Training and selection use
a zero-filled CG start. The same \method~architecture, hard data consistency
operator, and metric implementation are used for both benchmarks. The source
code and implementation details are available at
\url{https://github.com/yqx7150/HSSRecon}.

\textbf{\textit{Evaluation metrics:}}
All outputs are converted to root-sum-of-squares (RSS) magnitude images and
normalized by the maximum intensity of the fully sampled reference. We report
peak signal-to-noise ratio (PSNR, dB), structural similarity index measure
(SSIM), and normalized mean squared error (NMSE) with the same implementation
for every method.

\subsection{Experimental Results}
\label{subsec:experimental-results}
\textbf{\textit{Quantitative comparison:}}
Tables~\ref{tab:fastmri-brain-mask-r}--\ref{tab:stanford-bone-mask}
summarize the quantitative results under the two evaluation protocols.
Table~\ref{tab:fastmri-brain-mask-r} groups the fastMRI cases by sampling
pattern and acceleration factor and pools the three contrasts.
Table~\ref{tab:fastmri-brain-contrast-r} groups the cases by contrast and
acceleration factor and averages the three masks within each subject.
Table~\ref{tab:stanford-bone-mask} reports the three sampling patterns on the
Stanford cohort at $R=4$. Each entry gives the mean PSNR/SSIM/NMSE. The
rightmost columns in the two fastMRI tables report the raw two-sided paired
$t$-test on PSNR after averaging the six displayed mask--acceleration
combinations at $R=6,8$ for each subject within each contrast. The Stanford panel is
reported
descriptively because it serves as external validation and is scored on its
native full-FOV grid. Only $R=6$ and $R=8$ are shown for fastMRI. Table~I contains $n=15$
subject means for each mask and acceleration combination, Table~II contains
$n=5$ subject means for each contrast and acceleration combination, and Table~III
contains $n=10$ subject
means per sampling pattern.

\begin{table*}[!t]
\centering
\caption{Quantitative comparison on the fastMRI brain benchmark at $R=6$ and $R=8$. Entries are mean PSNR/SSIM/NMSE; the rightmost $p$-value is a raw two-sided paired $t$-test on PSNR after each contrast--subject unit is averaged over the six displayed mask--acceleration conditions at $R=6,8$.}
\label{tab:fastmri-brain-mask-r}
\scriptsize
\setlength{\tabcolsep}{3pt}
\renewcommand{\arraystretch}{1.16}
\resizebox{\textwidth}{!}{%
\begin{tabular}{l|*{6}{c}c}
\specialrule{\heavyrulewidth}{0pt}{0pt}
\rowcolor{tableheader}
\textbf{Method} & Poisson R6 & Poisson R8 & Radial R6 & Radial R8 & Uniform R6 & Uniform R8 & $p$-value \\
\specialrule{\lightrulewidth}{0pt}{0pt}
ZF & 28.18/0.7953/0.0281 & 27.84/0.7763/0.0308 & 28.52/0.7826/0.0266 & 26.71/0.7399/0.0394 & 23.95/0.6529/0.0715 & 23.50/0.6153/0.0791 & $3.4\times10^{-9}$ \\
SAKE~\cite{shin2014sake} & 31.56/0.7680/0.0145 & 30.74/0.7348/0.0179 & 31.13/0.7663/0.0149 & 29.82/0.7313/0.0197 & 28.46/0.7150/0.0251 & 26.84/0.6682/0.0361 & $2.7\times10^{-7}$ \\
ALOHA~\cite{jin2016hankel} & 31.78/0.8004/0.0143 & 30.86/0.7724/0.0176 & 31.21/0.8076/0.0150 & 29.62/0.7677/0.0210 & 27.68/0.7572/0.0316 & 26.34/0.7090/0.0425 & $5.9\times10^{-6}$ \\
SSDU~\cite{yaman2020ssdu} & 31.66/0.7933/0.0167 & 31.01/0.7660/0.0191 & 31.20/0.7925/0.0171 & 30.26/0.7624/0.0207 & 29.49/0.7615/0.0237 & 26.70/0.6883/0.0394 & $1.7\times10^{-3}$ \\
BOINR~\cite{yu2026bilevelinr} & 30.92/0.7808/0.0182 & 30.09/0.7417/0.0216 & 30.16/0.7610/0.0198 & 29.02/0.7207/0.0250 & 29.60/0.7467/0.0230 & 27.23/0.6738/0.0356 & $5.6\times10^{-8}$ \\
\method & \textbf{32.92}/\textbf{0.8367}/\textbf{0.0116} & \textbf{31.98}/\textbf{0.8099}/\textbf{0.0143} & \textbf{31.94}/\textbf{0.8251}/\textbf{0.0128} & \textbf{30.46}/\textbf{0.7915}/\textbf{0.0175} & \textbf{30.33}/\textbf{0.7929}/\textbf{0.0180} & \textbf{27.94}/\textbf{0.7419}/\textbf{0.0296} & -- \\
\bottomrule
\end{tabular}%
}
\end{table*}

\begin{table*}[!t]
\centering
\caption{Quantitative comparison on the fastMRI brain benchmark at $R=6$ and $R=8$. Entries are mean PSNR/SSIM/NMSE; the rightmost $p$-value is a raw two-sided paired $t$-test on PSNR after each contrast--subject unit is averaged over the six displayed mask--acceleration conditions at $R=6,8$.}
\label{tab:fastmri-brain-contrast-r}
\scriptsize
\setlength{\tabcolsep}{3pt}
\renewcommand{\arraystretch}{1.16}
\resizebox{\textwidth}{!}{%
\begin{tabular}{l|*{6}{c}c}
\specialrule{\heavyrulewidth}{0pt}{0pt}
\rowcolor{tableheader}
\textbf{Method} & AXFLAIR R6 & AXFLAIR R8 & AXT1PRE R6 & AXT1PRE R8 & AXT2 R6 & AXT2 R8 & $p$-value \\
\specialrule{\lightrulewidth}{0pt}{0pt}
ZF & 25.96/0.6867/0.0464 & 25.21/0.6464/0.0541 & 27.69/0.7616/0.0361 & 26.83/0.7380/0.0421 & 27.00/0.7824/0.0437 & 26.02/0.7471/0.0531 & $3.4\times10^{-9}$ \\
SAKE~\cite{shin2014sake} & 28.73/0.6927/0.0244 & 27.48/0.6523/0.0321 & 30.54/0.7363/0.0165 & 29.52/0.7036/0.0214 & 31.88/0.8203/0.0137 & 30.40/0.7784/0.0202 & $2.7\times10^{-7}$ \\
ALOHA~\cite{jin2016hankel} & 28.65/0.7422/0.0264 & 27.44/0.7003/0.0344 & 30.92/0.7847/0.0158 & 29.78/0.7493/0.0208 & 31.11/0.8383/0.0187 & 29.60/0.7995/0.0260 & $5.9\times10^{-6}$ \\
SSDU~\cite{yaman2020ssdu} & 28.02/0.7335/0.0325 & 26.82/0.6810/0.0421 & 30.88/0.7668/0.0162 & 29.64/0.7316/0.0225 & 33.45/0.8470/0.0088 & \textbf{31.52}/0.8041/\textbf{0.0146} & $1.7\times10^{-3}$ \\
BOINR~\cite{yu2026bilevelinr} & 27.68/0.7148/0.0337 & 26.56/0.6614/0.0424 & 30.59/0.7524/0.0169 & 29.43/0.7050/0.0222 & 32.41/0.8212/0.0104 & 30.35/0.7698/0.0178 & $5.6\times10^{-8}$ \\
\method & \textbf{29.56}/\textbf{0.7680}/\textbf{0.0222} & \textbf{28.10}/\textbf{0.7275}/\textbf{0.0302} & \textbf{32.10}/\textbf{0.8077}/\textbf{0.0116} & \textbf{30.85}/\textbf{0.7756}/\textbf{0.0155} & \textbf{33.53}/\textbf{0.8790}/\textbf{0.0085} & 31.44/\textbf{0.8403}/0.0157 & -- \\
\bottomrule
\end{tabular}%
}
\end{table*}

In Table~\ref{tab:fastmri-brain-mask-r}, \method~has the highest PSNR and SSIM
and the lowest NMSE in all six displayed $R=6/R=8$ conditions. Table~\ref{tab:fastmri-brain-contrast-r}
shows the same ordering for AXFLAIR and AXT1PRE. For AXT2, \method~leads at
$R=6$. At $R=8$, SSDU has a slightly higher PSNR (31.52 versus 31.44 dB) and
lower NMSE (0.0146 versus 0.0157), whereas \method~retains the higher SSIM
(0.8403 versus 0.8041). Relative to ZF and SAKE, \method~improves all six
displayed mask and acceleration combinations. The paired PSNR tests use one
observation for each subject within each contrast after averaging its six
displayed mask--acceleration combinations at $R=6,8$. The raw paired
$p$-values are below $10^{-4}$ for ZF, SAKE, ALOHA, and BOINR, and equal to
$1.7\times10^{-3}$ for SSDU.

Table~\ref{tab:stanford-bone-mask} shows that \method~achieves the highest
PSNR and SSIM under Poisson sampling and the highest PSNR under radial
sampling. Under uniform sampling, BOINR has the highest PSNR, while SSDU has
the highest SSIM and lowest NMSE. Across the three sampling patterns,
\method~achieves the lowest mean NMSE, although its advantage does not extend
to every reported metric.

\begin{table}[!t]
\centering
\caption{Quantitative comparison on the Stanford 2D FSE Bone10 panel at $R=4$. Entries are mean PSNR/SSIM/NMSE over 10 subjects and three interior slices per mask.}
\label{tab:stanford-bone-mask}
\scriptsize
\setlength{\tabcolsep}{1.8pt}
\renewcommand{\arraystretch}{1.08}
\resizebox{\columnwidth}{!}{%
\begin{tabular}{l|ccc}
\specialrule{\heavyrulewidth}{0pt}{0pt}
\rowcolor{tableheader}
\textbf{Method} & Poisson R4 & Radial R4 & Uniform R4 \\
\specialrule{\lightrulewidth}{0pt}{0pt}
ZF & 32.36/0.8439/0.0304 & 30.22/0.7940/0.0409 & 28.10/0.6910/0.0606 \\
SAKE~\cite{shin2014sake} & 36.21/0.8911/0.0187 & 33.60/0.8577/0.0230 & 30.78/0.7619/0.0381 \\
ALOHA~\cite{jin2016hankel} & 35.85/0.9031/0.0194 & 32.57/0.8619/0.0273 & 29.57/0.7480/0.0458 \\
SSDU~\cite{yaman2020ssdu} & 36.14/0.9072/0.0197 & 34.07/\textbf{0.8807}/0.0258 & 32.58/\textbf{0.8245}/\textbf{0.0334} \\
BOINR~\cite{yu2026bilevelinr} & 36.07/0.9035/0.0187 & 34.07/0.8608/0.0231 & \textbf{32.62}/0.7928/0.0350 \\
\method & \textbf{37.15}/\textbf{0.9199}/\textbf{0.0185} & \textbf{34.30}/0.8790/\textbf{0.0221} & 31.29/0.7783/0.0344 \\
\bottomrule
\end{tabular}%
}
\end{table}

\begin{figure*}[!t]
    \centering
    \IfFileExists{figures/fig-fastmri-brain-main-comparison.pdf}{%
        \includegraphics[width=\textwidth]{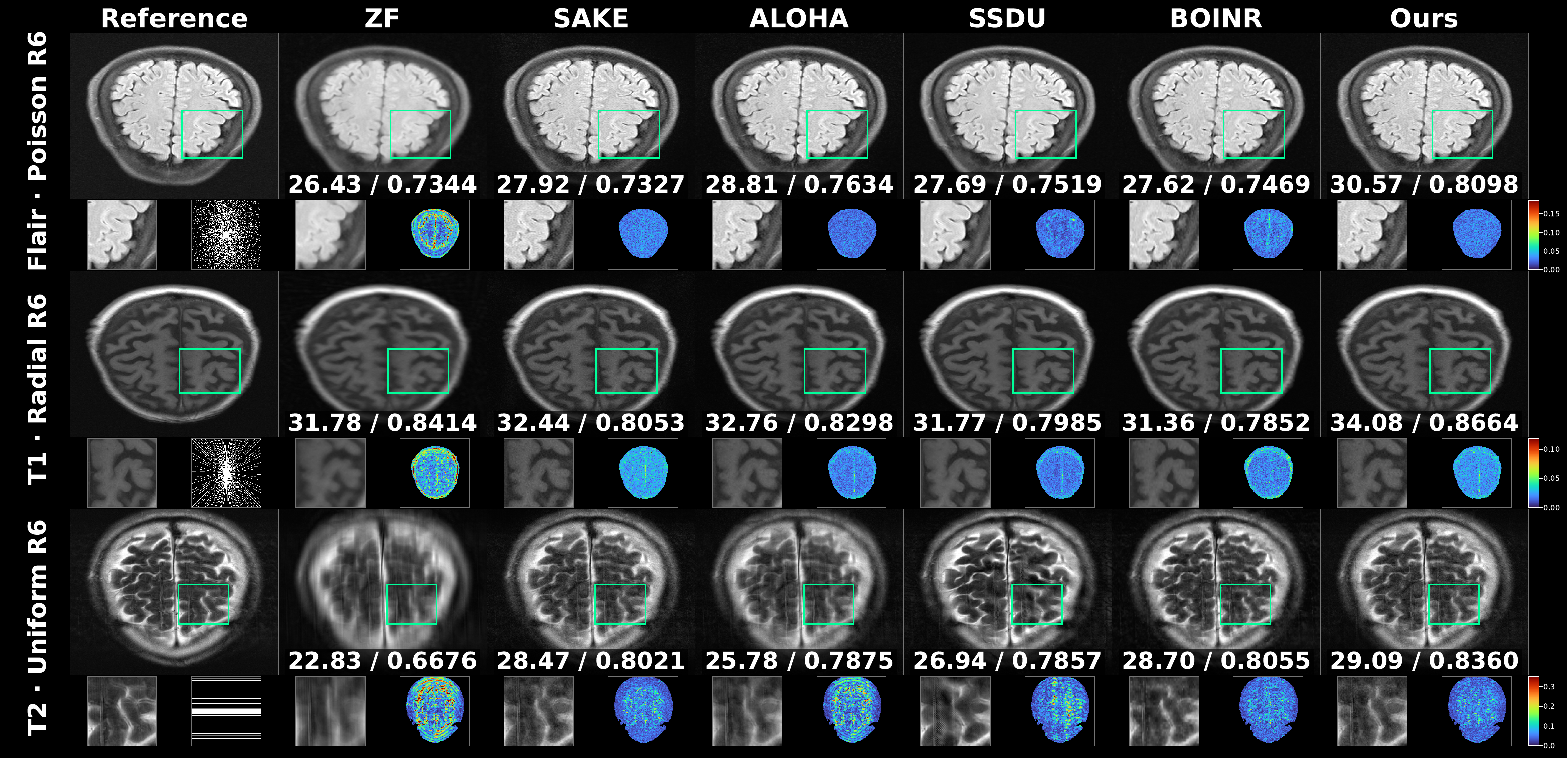}%
    }{%
        \IfFileExists{figures/fig-fastmri-brain-main-comparison.png}{%
            \includegraphics[width=\textwidth]{figures/fig-fastmri-brain-main-comparison.png}%
        }{%
            \placeholderfigure{1.55in}{Main fastMRI brain comparison: qualitative
            examples across the three masks and acceleration rates,
            with RSS reconstructions, ROI/error details, and compact PSNR/SSIM
            annotations.}%
        }%
    }
    \caption{Qualitative fastMRI brain comparisons at $R=6$. The rows labeled
    Flair, T1, and T2 in the image correspond to AXFLAIR/Poisson,
    AXT1PRE/radial, and AXT2/uniform. The image label ``Radial'' denotes
    the rasterized Cartesian-grid mask. Columns show the reference, ZF, SAKE,
    ALOHA, SSDU, BOINR, and \method~(labeled ``Ours'' in the image). Each
    reconstruction is annotated
    with PSNR/SSIM from the same scoring procedure. Below the reference are the
    ROI detail and acquisition mask. Below each reconstruction are the
    corresponding ROI detail and absolute RSS error map over the brain
    foreground. A common reconstruction window is used, with a shared error
    scale within each row.}
\label{fig:fastmri-brain-main}
\end{figure*}

\begin{figure*}[!t]
    \centering
    \IfFileExists{figures/fig-stanford-bone-main-comparison.pdf}{%
        \includegraphics[width=\textwidth]{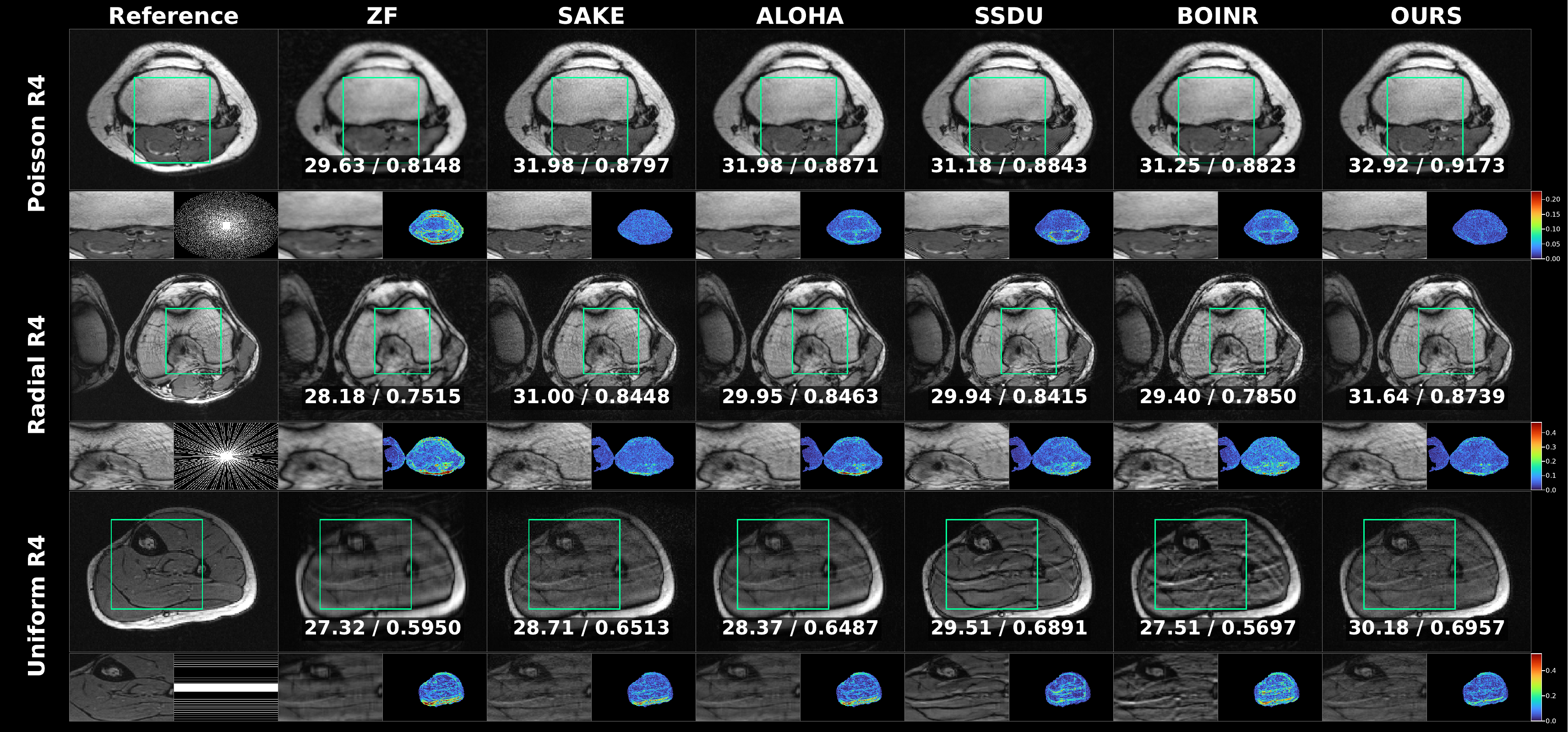}%
    }{%
        \IfFileExists{figures/fig-stanford-bone-main-comparison.png}{%
            \includegraphics[width=\textwidth]{figures/fig-stanford-bone-main-comparison.png}%
        }{%
            \placeholderfigure{1.55in}{Stanford 2D FSE Bone10 qualitative
            comparison across Poisson, radial, and uniform sampling.}%
        }%
    }
    \caption{Qualitative Stanford 2D FSE Bone10 comparisons at $R=4$.
    Rows correspond to Poisson, radial, and uniform sampling. Columns show the
    reference, ZF, SAKE, ALOHA, SSDU, BOINR, and \method. The lower panels show
    the acquired mask for the reference and foreground absolute RSS error maps
    for the reconstructions. Scores use the original full-FOV image grid.}
    \label{fig:stanford-bone-main}
\end{figure*}

\textbf{\textit{Qualitative comparison:}}
Fig.~\ref{fig:fastmri-brain-main} presents three fastMRI brain examples at
$R=6$ from different contrasts and sampling patterns. Fig.~\ref{fig:stanford-bone-main}
presents one Stanford example for each sampling pattern at $R=4$. In these
examples, \method~reduces structured aliasing and local boundary errors. The
uniform Stanford example also illustrates the metric differences reported in
Table~\ref{tab:stanford-bone-mask}.

\subsection{Ablation Study}
\label{subsec:ablation-results}
We evaluate the contribution of each component on nine AXT2/Poisson $R=6$
slices, with three per subject. Apart
from the tested change, learned variants share the acquired data, masks,
$3\times3$ Hankel lift, two views, 24 CG steps, and hard data consistency.
The training budget for this analysis is capped at 600 updates.
Checkpoint selection uses $\Gamma$, excluded from gradient updates, before
reference-based scoring. Table~\ref{tab:ablation} gives mean PSNR, SSIM, and
NMSE across the nine slices.

\Needspace{5\baselineskip}
For the Fixed operator control, we construct a Hankel matrix from the full
acquired k-space, with unacquired entries set to zero. The 16 lowest-energy
right singular vectors define the structural subspace, and the next eight
define the innovation subspace. Its innovation weight is set to match the
learned operator's trace without reference data. We compute the fixed operator
once per scan and hold it fixed across both views
and all reconstruction solves. Both settings use the same CG solver and hard
data consistency. Relative to the fixed operator, the learned operator
improves PSNR by 6.23 dB. This comparison assesses
learning and view conditioning jointly.

The physical split removes all duplicate copies of the target coordinates from
the input. In contrast, splitting after the Hankel lift left 99.996\% of the
target coordinates accessible through duplicate entries. The leakage control
therefore reached a near-zero first-step training loss of
$2.68\times10^{-8}$, compared with 0.118 for the physical split. However, its
held-out $\Gamma$ loss at the first evaluation was 0.0793, compared with 0.0172,
and its reconstruction PSNR was 4.73 dB lower.

\begin{table}[!t]
\centering
\caption{Subspace and split ablations on nine AXT2/Poisson $R=6$ slices.}
\label{tab:ablation}
\footnotesize
\setlength{\tabcolsep}{2.2pt}
\renewcommand{\arraystretch}{1.05}
\begin{tabularx}{\columnwidth}{l|*{3}{>{\centering\arraybackslash}X}}
\specialrule{\heavyrulewidth}{0pt}{0pt}
\rowcolor{tableheader}
\textbf{Setting} & PSNR (dB) & SSIM &
NMSE \\
\specialrule{\lightrulewidth}{0pt}{0pt}
\method                     & 35.12 & 0.9365 & 0.0055 \\
Fixed operator              & 28.89 & 0.8304 & 0.0243 \\
Hankel-entry split         & 30.39 & 0.8583 & 0.0169 \\
\bottomrule
\end{tabularx}%
\end{table}

\begin{figure}[!t]
    \centering
    \includegraphics[width=0.96\columnwidth]{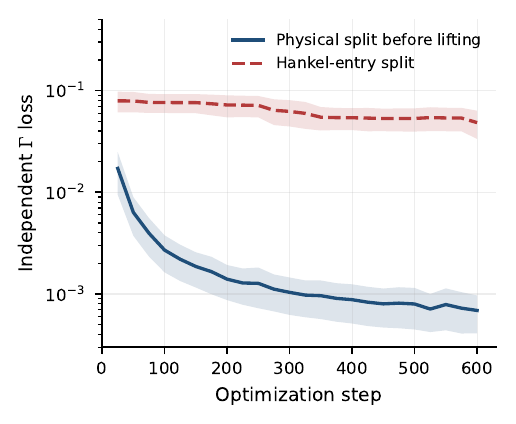}
    \caption{Held-out $\Gamma$ selection loss for the physical split
    before lifting and the split of Hankel entries after lifting. The curves show
    means over nine slices. Shaded bands indicate one sample
    standard deviation. The vertical axis is logarithmic. $\Gamma$ is excluded from gradient
    updates but used for checkpoint selection. The split after lifting retains
    higher selection loss despite its near-zero training loss.}
    \label{fig:ablation-leakage-loss}
\end{figure}

\section{Discussion}
Compared with physical splitting, the control that splits after lifting reached
near-zero
training loss but had higher $\Gamma$ selection loss and poorer reconstruction.
Nearly every held-out
physical coordinate was also present in the input after lifting, consistent
with prediction from duplicate entries. Splitting physical groups before
lifting removes this route. Multiplicity correction prevents window overlap
from changing the weight of a physical coefficient.

The quantitative results support the value of this framework under the
prespecified fastMRI brain evaluation setting. In the displayed summary over
mask and acceleration combinations, HSSRecon
has the highest mean PSNR and SSIM in all six $R=6/R=8$ groups, and it improves
over ZF and SAKE throughout the six groups in Table~\ref{tab:fastmri-brain-mask-r}.
Table~\ref{tab:fastmri-brain-contrast-r} gives a more qualified view.
The gains are clear for AXFLAIR and AXT1PRE, whereas the AXT2 R8 PSNR is close
to, and slightly below, SSDU. HSSRecon still has the higher AXT2 R8 SSIM in
that comparison. These results indicate that the learned Hankel operator is
useful beyond a single contrast or sampling pattern.

The independent Stanford 2D FSE panel provides a complementary robustness
check. The ordering is favorable for \method~under Poisson and radial
sampling, whereas the uniform results favor BOINR or SSDU on individual
metrics. This pattern is consistent with the role of the learned Hankel
operator: its benefit depends on whether the sampling geometry supplies local
multi-coil relations that can be identified from the acquired data.
The larger relative gains on fastMRI brain than on Stanford Bone10 may reflect
the amount of structural evidence available to the scan-specific learner. The
fastMRI carrier uses a larger, standardized working grid and 16 or 20 coils,
whereas the Stanford carriers have smaller, variable matrices with 16 coils.
The fastMRI setup provides more overlapping Hankel neighborhoods and, in some
cases, richer inter-coil measurements for subspace estimation. The fastMRI results
also use $R=6/8$, whereas Stanford uses $R=4$, so the two panels do not have
the same reconstruction difficulty or improvement headroom.

The separation between the learner and the solver also clarifies the role of
each part of the framework. The encoder does not output the missing coefficients.
It produces factor matrices whose QR projectors define the learned Hankel
penalty operator. The constrained CG procedure then determines the physical
$k$-space values while preserving the acquired samples. This separation makes
the projector, the hard data-consistency residual, and the solver residual
available for separate inspection. The theoretical result guarantees a
well-posed quadratic problem and a standard finite-step CG bound under the
stated conditions.

The present evaluation covers only two 2D cohorts and does not isolate
frequency-specific fidelity. Future work will extend validation to larger
subject-matched 3D and dynamic MRI cohorts, where inter-slice and temporal
dependencies may provide additional constraints, and will use frequency-resolved
metrics to assess high-frequency fidelity~\cite{peng2026threedimensional,
shen2026imjplus,xu2025ssflrecon}.

\section{Conclusion}
In this paper, we proposed HSSRecon, a scan-specific Hankel-subspace
self-supervised reconstruction framework that combines physical-group
self-supervision with a multiplicity-aware Hankel penalty and
reconstruction with hard data consistency. Its analysis
establishes an exact loss correspondence and a well-posed constrained solve
under the stated conditions. The tested fastMRI brain results favor \method~in
pooled summaries over mask and acceleration combinations, with a contrast-specific
exception at AXT2
$R=8$.
On the independent Stanford Bone10 panel, HSSRecon achieved the highest PSNR
under Poisson and radial sampling, while other methods led under uniform
sampling. The ablations further show that physical splitting prevents leakage
through duplicate Hankel entries and that the learned operator improves PSNR
over the fixed-operator control.

\appendix

\subsection{Proof of Proposition~\ref{prop:quotient}}

For a pure lift that copies each physical entry, each lifted entry is a copy of
exactly one physical
coordinate. Consequently, $\mathcal{H}^{*}\mathcal{H}=D$ is diagonal and its
$i$th diagonal element is the number $d_i$ of copies of coordinate $i$.
Because $P_{\mathcal{H}}Z$ belongs to the range of $\mathcal{H}$, all copies
associated with coordinate $i$ have the same value,
$[\mathcal{H}^{\dagger}P_{\mathcal{H}}Z]_i$. Summing the $d_i$ identical
terms in Eq.~\eqref{eq:quotient_identity} cancels the factor $d_i$ and
gives the corresponding physical-coordinate loss divided by $r_i$.

If the split is performed before lifting, $y_{\Theta_v}$ is zero on every
physical coordinate in $\Lambda_v$. Since $\mathcal{H}$ only copies physical
coordinates, every lifted copy of a held-out coordinate is also absent from
$\mathcal{H}(y_{\Theta_v})$. This proves both the exact quotient identity and
the exclusion of duplicate held-out entries at the level of the physical support.
Finally, the normalization gives:
\begin{equation}
\overline{\mathcal H}^{*}\overline{\mathcal H}
=D^{-1/2}\mathcal H^{*}\mathcal H D^{-1/2}=I.
\label{eq:app-lift-isometry}
\end{equation}
This proves Eq.~\eqref{eq:lift_isometry}.\hfill$\square$

\subsection{Proof of Theorem~\ref{thm:solver}}

Since $W_{\theta,v}\succeq0$, for any vector $u$:
\begin{equation}
u^{H}R_{\theta,v}u
=\lambda_H\| (\overline{\mathcal{H}}u)
W_{\theta,v}^{1/2}\|_F^2+\lambda_0\|u\|_2^2
\geq\lambda_0\|u\|_2^2.
\label{eq:app-positive-definiteness}
\end{equation}
Eq.~\eqref{eq:lift_isometry} and
$\|ZW_{\theta,v}^{1/2}\|_F^2\leq w_v\|Z\|_F^2$ also give
$u^HR_{\theta,v}u\leq(\lambda_0+\lambda_Hw_v)\|u\|_2^2$.
Since $N_v$ is orthonormal, both bounds hold for $A_v$, establishing
Eq.~\eqref{eq:solver_spectrum}. Replacing the normalized lift by
$\mathcal H$ under the same $W_{\theta,v}$ and regularization weights gives
the upper spectral bound $\lambda_0+\lambda_Hw_v\max_i d_i$, since
$\|\mathcal H\|_2^2=\max_i d_i$. For full-rank QR factors,
$P_{s,v}P_{d,v}=0$. The structural subspace, innovation subspace, and their
orthogonal complement are invariant spaces of $W_{\theta,v}$ with
eigenvalues $1$, $\alpha_v$, and $0$, respectively.
Thus $A_v$ is Hermitian positive definite, $\mu_v\geq\lambda_0$, and the
reduced quadratic has the unique solution $z_\infty=A_v^{-1}c_v$.
Since $P_{\Theta_v}N_v=0$, every iterate
$k_{\mathrm{dc},v}+N_vz_j$ retains the input measurements.

For $e_K=z_K-z_\infty$, the standard exact-arithmetic CG bound is:
\begin{equation}
\|e_K\|_{A_v}\leq
2\left(\frac{\sqrt{\kappa_v}-1}{\sqrt{\kappa_v}+1}\right)^K
\|e_0\|_{A_v}.
\label{eq:app-cg-energy-bound}
\end{equation}
Because $N_v$ is orthonormal and $A_v\succeq\mu_vI$,
$\delta_{\mathrm{CG}}(K)=\|N_ve_K\|_2=\|e_K\|_2
\leq\|e_K\|_{A_v}/\sqrt{\mu_v}$, proving Eq.~\eqref{eq:cg_bound}.
Also, $r_K=A_v(z_\infty-z_K)$, so
$\delta_{\mathrm{CG}}(K)=\|A_v^{-1}r_K\|_2
\leq\|r_K\|_2/\mu_v\leq\|r_K\|_2/\lambda_0$, proving
Eq.~\eqref{eq:cg_residual_bound}. The residual bound holds for any feasible
iterate with this exact reduced-system residual.\hfill$\square$

\subsection{Proof of Corollary~\ref{cor:reconstruction-error}}

The chosen $k_{\mathrm{ref}}$ satisfies
$P_{\Theta_v}k_{\mathrm{ref}}=P_{\Theta_v}y$ and
$k_{\mathrm{ref}}-k_\star=P_{\Theta_v}(y-k_\star)$.
Let $d=\widehat{k}_{\theta,v}^{(\infty)}-k_{\mathrm{ref}}=N_vu$.
The free-space first-order condition gives:
\begin{equation}
0=A_vu+\lambda_H g_{H,v}(k_{\mathrm{ref}})
  +\lambda_0N_v^Hk_{\mathrm{ref}}.
\label{eq:app-free-optimality}
\end{equation}
Solving this first-order condition for $d$ separates the Hankel and ridge
contributions:
\begin{equation}
d=-N_vA_v^{-1}\bigl[\lambda_Hg_{H,v}(k_{\mathrm{ref}})
   +\lambda_0N_v^Hk_{\mathrm{ref}}\bigr].
\label{eq:app-regularization-deviation}
\end{equation}
Using $\|A_v^{-1}\|_2=1/\mu_v$ yields:
\begin{equation}
    \|d\|_2\leq
    \frac{\lambda_H}{\mu_v}\|g_{H,v}(k_{\mathrm{ref}})\|_2
    +\frac{\lambda_0}{\mu_v}\|N_v^Hk_{\mathrm{ref}}\|_2.
    \label{eq:mismatch_bound}
\end{equation}
The triangle inequality, together with the identity for the comparison point and
the definition of $\delta_{\mathrm{CG}}(K)$, gives
Eq.~\eqref{eq:total_bound}. Moreover, hard data consistency gives
$P_{\Theta_v}(\widehat{k}_{\theta,v}^{(K)}-k_\star)
=P_{\Theta_v}(y-k_\star)$. Since a coordinate projection cannot increase
the norm:
\begin{equation}
\|\widehat{k}_{\theta,v}^{(K)}-k_\star\|_2
\geq\|P_{\Theta_v}(y-k_\star)\|_2.
\label{eq:app-measurement-error-floor}
\end{equation}
The input measurement discrepancy is therefore unavoidable.\hfill$\square$

\bibliographystyle{ieeetr}
\bibliography{references}

@article{pruessmann1999sense,
  title={{SENSE}: sensitivity encoding for fast {MRI}},
  author={Pruessmann, Klaas P. and Weiger, Markus and Scheidegger, Markus B. and Boesiger, Peter},
  journal={Magnetic Resonance in Medicine},
  volume={42},
  number={5},
  pages={952--962},
  year={1999},
  publisher={Wiley}
}

@article{griswold2002grappa,
  title={Generalized autocalibrating partially parallel acquisitions ({GRAPPA})},
  author={Griswold, Mark A. and Jakob, Peter M. and Heidemann, Robin M. and Nittka, Mathias and Jellus, Vladimir and Wang, Jianmin and Kiefer, Berthold and Haase, Axel},
  journal={Magnetic Resonance in Medicine},
  volume={47},
  number={6},
  pages={1202--1210},
  year={2002},
  publisher={Wiley}
}

@article{lustig2010spirit,
  title={{SPIRiT}: Iterative self-consistent parallel imaging reconstruction from arbitrary k-space},
  author={Lustig, Michael and Pauly, John M.},
  journal={Magnetic Resonance in Medicine},
  volume={64},
  number={2},
  pages={457--471},
  year={2010},
  publisher={Wiley}
}

@article{shin2014sake,
  title={Calibrationless parallel imaging reconstruction based on structured low-rank matrix completion},
  author={Shin, Peter J. and Larson, Peder E. Z. and Ohliger, Michael A. and Elad, Michael and Pauly, John M. and Vigneron, Daniel B. and Lustig, Michael},
  journal={Magnetic Resonance in Medicine},
  volume={72},
  number={4},
  pages={959--970},
  year={2014},
  publisher={Wiley}
}

@article{haldar2014loraks,
  title={Low-rank modeling of local k-space neighborhoods ({LORAKS}) for constrained {MRI}},
  author={Haldar, Justin P.},
  journal={IEEE Transactions on Medical Imaging},
  volume={33},
  number={3},
  pages={668--681},
  year={2014},
  publisher={IEEE}
}

@article{haldar2016ploraks,
  title={{P-LORAKS}: Low-rank modeling of local k-space neighborhoods with parallel imaging data},
  author={Haldar, Justin P. and Zhuo, Jingwei},
  journal={Magnetic Resonance in Medicine},
  volume={75},
  number={4},
  pages={1499--1514},
  year={2016},
  doi={10.1002/mrm.25717},
  publisher={Wiley}
}

@article{yu2026bilevelinr,
  title={Bilevel Optimized Implicit Neural Representation for Scan-Specific Accelerated {MRI} Reconstruction},
  author={Yu, Hongze and Fessler, Jeffrey A. and Jiang, Yun},
  journal={IEEE Transactions on Medical Imaging},
  volume={45},
  number={7},
  pages={3792--3808},
  year={2026},
  doi={10.1109/TMI.2026.3686724},
  publisher={IEEE}
}

@article{jin2016hankel,
  title={A general framework for compressed sensing and parallel {MRI} using annihilating filter based low-rank {Hankel} matrix},
  author={Jin, Kyong Hwan and Lee, Dongwook and Ye, Jong Chul},
  journal={IEEE Transactions on Computational Imaging},
  volume={2},
  number={4},
  pages={480--495},
  year={2016},
  doi={10.1109/TCI.2016.2601296},
  publisher={IEEE}
}

@article{park2017smshsl,
  title={{SMS-HSL}: Simultaneous multislice aliasing separation exploiting {Hankel} subspace learning},
  author={Park, Suhyung and Park, Jaeseok},
  journal={Magnetic Resonance in Medicine},
  volume={78},
  number={4},
  pages={1392--1404},
  year={2017},
  doi={10.1002/mrm.26527},
  publisher={Wiley}
}

@article{kim2021smst1,
  title={Simultaneous Multislice Brain {MRI} {T1} Mapping with Improved Low-Rank Modeling},
  author={Kim, Sugil and Park, Suhyung},
  journal={Tomography},
  volume={7},
  number={4},
  pages={545--554},
  year={2021},
  doi={10.3390/tomography7040047},
  publisher={MDPI}
}

@article{kim2023modelsmshsl,
  title={Model-Based Simultaneous Multi-Slice ({SMS}) Reconstruction with {Hankel} Subspace Learning for Accelerated {MR} {T1} Mapping},
  author={Kim, Sugil and Wu, Hua and Han, Jae-Ho},
  journal={Mathematics},
  volume={11},
  number={13},
  pages={2963},
  year={2023},
  doi={10.3390/math11132963},
  publisher={MDPI}
}

@article{hammernik2018variational,
  title={Learning a variational network for reconstruction of accelerated {MRI} data},
  author={Hammernik, Kerstin and Klatzer, Teresa and Kobler, Erich and Recht, Michael P. and Sodickson, Daniel K. and Pock, Thomas and Knoll, Florian},
  journal={Magnetic Resonance in Medicine},
  volume={79},
  number={6},
  pages={3055--3071},
  year={2018},
  publisher={Wiley}
}

@article{aggarwal2019modl,
  title={{MoDL}: Model-based deep learning architecture for inverse problems},
  author={Aggarwal, Hemant K. and Mani, Merry P. and Jacob, Mathews},
  journal={IEEE Transactions on Medical Imaging},
  volume={38},
  number={2},
  pages={394--405},
  year={2019},
  publisher={IEEE}
}

@article{akcakaya2019raki,
  title={Scan-specific robust artificial-neural-networks for k-space interpolation ({RAKI}) reconstruction: Database-free deep learning for fast imaging},
  author={Ak{\c{c}}akaya, Mehmet and Moeller, Steen and Weing{\"a}rtner, Sebastian and U{\u{g}}urbil, Kamil},
  journal={Magnetic Resonance in Medicine},
  volume={81},
  number={1},
  pages={439--453},
  year={2019},
  publisher={Wiley}
}

@article{yaman2020ssdu,
  title={Self-supervised learning of physics-guided reconstruction neural networks without fully sampled reference data},
  author={Yaman, Burhaneddin and Hosseini, Seyed Amir Hossein and Moeller, Steen and Ellermann, Jutta and U{\u{g}}urbil, Kamil and Ak{\c{c}}akaya, Mehmet},
  journal={Magnetic Resonance in Medicine},
  volume={84},
  number={6},
  pages={3172--3191},
  year={2020},
  publisher={Wiley}
}

@article{knoll2020fastmri,
  title={{fastMRI}: A publicly available raw k-space and {DICOM} dataset of knee images for accelerated {MR} image reconstruction using machine learning},
  author={Knoll, Florian and Zbontar, Jure and Sriram, Anuroop and Muckley, Matthew J. and Bruno, Mary and Defazio, Aaron and Parente, Marc and Geras, Krzysztof J. and Katsnelson, Joe and Chandarana, Hersh and Zhang, Zizhao and Drozdzal, Michal and Romero, Adriana and Rabbat, Michael and Vincent, Pascal and Pinkerton, James and Wang, Duo and Yakubova, Nafissa and Owens, Erich and Zitnick, C. Lawrence and Recht, Michael P. and Sodickson, Daniel K. and Lui, Yvonne W.},
  journal={Radiology: Artificial Intelligence},
  volume={2},
  number={1},
  pages={e190007},
  year={2020},
  doi={10.1148/ryai.2020190007},
  publisher={Radiological Society of North America}
}

@article{muckley2021fastmriChallenge,
  title={Results of the 2020 {fastMRI} Challenge for Machine Learning {MR} Image Reconstruction},
  author={Muckley, Matthew J. and Riemenschneider, Bruno and Radmanesh, Alireza and Kim, Sunwoo and Jeong, Geunu and Ko, Jingyu and Jun, Yohan and Shin, Hyungseob and Hwang, Dosik and Mostapha, Mahmoud and Arberet, Simon and Nickel, Dominik and Ramzi, Zaccharie and Ciuciu, Philippe and Starck, Jean-Luc and Teuwen, Jonas and Karkalousos, Dimitrios and Zhang, Chaoping and Sriram, Anuroop and Huang, Zhengnan and Yakubova, Nafissa and Lui, Yvonne W. and Knoll, Florian},
  journal={IEEE Transactions on Medical Imaging},
  volume={40},
  number={9},
  pages={2306--2317},
  year={2021},
  doi={10.1109/TMI.2021.3075856},
  publisher={IEEE}
}

@misc{cheng2018stanford2dfse,
  author       = {Cheng, Joseph Y.},
  title        = {Stanford 2D FSE},
  year         = {2018},
  howpublished = {MRIdata.org},
  url          = {http://mridata.org/list?project=Stanford2DFSE},
  note         = {Accessed: Sep. 30, 2026}
}

@inproceedings{fabian2021mraugment,
  author    = {Fabian, Zalan and Heckel, Reinhard and Soltanolkotabi, Mahdi},
  title     = {Data augmentation for deep learning based accelerated {MRI} reconstruction with limited data},
  booktitle = {Proceedings of the 38th International Conference on Machine Learning},
  series    = {Proceedings of Machine Learning Research},
  volume    = {139},
  pages     = {3057--3067},
  year      = {2021},
  publisher = {PMLR},
  url       = {https://proceedings.mlr.press/v139/fabian21a.html}
}

@article{uecker2014espirit,
  title={{ESPIRiT}--an eigenvalue approach to autocalibrating parallel {MRI}: Where {SENSE} meets {GRAPPA}},
  author={Uecker, Martin and Lai, Peng and Murphy, Mark J. and Virtue, Patrick and Elad, Michael and Pauly, John M. and Vasanawala, Shreyas S. and Lustig, Michael},
  journal={Magnetic Resonance in Medicine},
  volume={71},
  number={3},
  pages={990--1001},
  year={2014},
  doi={10.1002/mrm.24751},
  publisher={Wiley}
}

@article{peng2026threedimensional,
  author  = {Peng, Tengya and Zha, Ruyi and Li, Zhen and Liu, Xiaofeng and Zou, Qing},
  title   = {Three-Dimensional {MRI} Reconstruction With {3D} Gaussian Representations: Tackling the Undersampling Problem},
  journal = {IEEE Transactions on Medical Imaging},
  year    = {2026},
  volume  = {45},
  number  = {5},
  pages   = {1905--1917},
  doi     = {10.1109/TMI.2025.3642134}
}

@article{shen2026imjplus,
  author  = {Shen, Xin and Feng, Jie and Li, Zhenghao and Zou, Qing and Zhang, Yuyao and Wei, Hongjiang},
  title   = {Free-Breathing Dynamic {MRI} Reconstruction via Joint Time-Dependent Coil Sensitivity Estimation Using Implicit Neural Representation},
  journal = {Medical Image Analysis},
  year    = {2026},
  volume  = {108},
  pages   = {103847},
  doi     = {10.1016/j.media.2025.103847}
}

@article{pramanik2020deepSLR,
  author  = {Pramanik, Aniket and Aggarwal, Hemant Kumar and Jacob, Mathews},
  title   = {Deep Generalization of Structured Low-Rank Algorithms ({Deep-SLR})},
  journal = {IEEE Transactions on Medical Imaging},
  year    = {2020},
  volume  = {39},
  number  = {12},
  pages   = {4186--4197},
  doi     = {10.1109/TMI.2020.3014581}
}

@article{peng2023hkgm,
  author  = {Peng, Hong and Jiang, Chen and Cheng, Jing and Zhang, Minghui and Wang, Shanshan and Liang, Dong and Liu, Qiegen},
  title   = {One-Shot Generative Prior in {Hankel}-$k$-space for Parallel Imaging Reconstruction},
  journal = {IEEE Transactions on Medical Imaging},
  year    = {2023},
  volume  = {42},
  number  = {11},
  pages   = {3420--3435},
  doi     = {10.1109/TMI.2023.3288219}
}

@article{luo2025hsspgd,
  author  = {Luo, Chen and Wang, Huayu and Liu, Yuanyuan and Xie, Taofeng and Chen, Guoqing and Jin, Qiyu and Liang, Dong and Cui, Zhuo-Xu},
  title   = {Matrix Completion-Informed Deep Unfolded Equilibrium Models for Self-Supervised $k$-Space Interpolation in {MRI}},
  journal = {Medical Physics},
  year    = {2025},
  volume  = {52},
  number  = {7},
  pages   = {e17924},
  doi     = {10.1002/mp.17924}
}

@article{yaman2022multimask,
  author  = {Yaman, Burhaneddin and Gu, Hongyi and Hosseini, Seyed Amir Hossein and Demirel, Omer Burak and Moeller, Steen and Ellermann, Jutta and U{\u{g}}urbil, Kamil and Ak{\c{c}}akaya, Mehmet},
  title   = {Multi-Mask Self-Supervised Learning for Physics-Guided Neural Networks in Highly Accelerated {MRI}},
  journal = {NMR in Biomedicine},
  year    = {2022},
  volume  = {35},
  number  = {12},
  pages   = {e4798},
  doi     = {10.1002/nbm.4798}
}

@inproceedings{yaman2021zsssl,
  author    = {Yaman, Burhaneddin and Hosseini, Seyed Amir Hossein and Akcakaya, Mehmet},
  title     = {Zero-Shot Self-Supervised Learning for {MRI} Reconstruction},
  booktitle = {International Conference on Learning Representations},
  year      = {2022},
  url       = {https://openreview.net/forum?id=085y6YPaYjP},
  note      = {[Online]. Available: \url{https://openreview.net/forum?id=085y6YPaYjP}}
}

@article{millard2022noisier2noise,
  author  = {Millard, Charles and Chiew, Mark},
  title   = {A Theoretical Framework for Self-Supervised {MR} Image Reconstruction Using Sub-Sampling via Variable Density {Noisier2Noise}},
  journal = {IEEE Transactions on Computational Imaging},
  volume  = {9},
  pages   = {707--720},
  year    = {2023},
  doi     = {10.1109/TCI.2023.3299212}
}

@misc{wang2024kband,
  author       = {Wang, Frederic and Qi, Han and De Goyeneche, Alfredo and Heckel, Reinhard and Lustig, Michael and Shimron, Efrat},
  title        = {{K-band}: Self-Supervised {MRI} Reconstruction via Stochastic Gradient Descent over $k$-space Subsets},
  year         = {2023},
  howpublished = {arXiv preprint arXiv:2308.02958},
  url          = {https://arxiv.org/abs/2308.02958},
  eprint       = {2308.02958},
  archivePrefix = {arXiv},
  primaryClass = {eess.IV},
  note         = {Revised 2024}
}

@inproceedings{batson2019noise2self,
  author    = {Batson, Joshua and Royer, Loic},
  title     = {{Noise2Self}: Blind Denoising by Self-Supervision},
  booktitle = {Proceedings of the 36th International Conference on Machine Learning},
  series    = {Proceedings of Machine Learning Research},
  volume    = {97},
  pages     = {524--533},
  year      = {2019},
  publisher = {PMLR},
  url       = {https://proceedings.mlr.press/v97/batson19a.html}
}

@inproceedings{joo2025aespa,
  author    = {Joo, Jinho and Kim, Hyeseong and Won, Hyeyeon and Lee, Deukhee and Eo, Taejoon and Hwang, Dosik},
  title     = {{AeSPa}: Attention-Guided Self-Supervised Parallel Imaging for {MRI} Reconstruction},
  booktitle = {Proceedings of the IEEE/CVF Conference on Computer Vision and Pattern Recognition},
  pages     = {5217--5226},
  year      = {2025},
  month     = {June},
  publisher = {IEEE/CVF},
  url       = {https://openaccess.thecvf.com/content/CVPR2025/html/Joo_AeSPa__Attention-guided_Self-supervised_Parallel_Imaging_for_MRI_Reconstruction_CVPR_2025_paper.html}
}

@misc{xu2026units,
  author       = {Xu, Siying and Hammernik, Kerstin and Rueckert, Daniel and Gatidis, Sergios and K{\"u}stner, Thomas},
  title        = {Towards a Unified Theoretical Framework for Splitting-Based Self-Supervised {MRI} Reconstruction},
  year         = {2026},
  howpublished = {arXiv preprint arXiv:2601.04775},
  note         = {Preprint, version 3 revised 2026-05-06},
  url          = {https://arxiv.org/abs/2601.04775},
  eprint       = {2601.04775},
  archivePrefix = {arXiv},
  primaryClass = {eess.IV}
}

@article{ye2018deepframelets,
  author    = {Ye, Jong Chul and Han, Yoseob and Cha, Eunju},
  title     = {Deep Convolutional Framelets: A General Deep Learning Framework for Inverse Problems},
  journal   = {SIAM Journal on Imaging Sciences},
  year      = {2018},
  volume    = {11},
  number    = {2},
  pages     = {991--1048},
  doi       = {10.1137/17M1141771},
  publisher = {SIAM}
}

@inproceedings{zhao2020convframework,
  author    = {Zhao, Shen and Potter, Lee C. and Lee, Kiryung and Ahmad, Rizwan},
  title     = {Convolutional Framework for Accelerated Magnetic Resonance Imaging},
  booktitle = {2020 IEEE 17th International Symposium on Biomedical Imaging},
  pages     = {1065--1068},
  year      = {2020},
  doi       = {10.1109/ISBI45749.2020.9098393},
  publisher = {IEEE}
}

@article{qian2025donate,
  author    = {Qian, Chen and Han, Mingyang and Zhu, Liuhong and Wang, Zi and Guan, Feiqiang and Guo, Yucheng and Ruan, Dan and Guo, Yi and Kang, Taishan and Lin, Jianzhong and Wang, Chengyan and Mani, Merry and Jacob, Mathews and Lin, Meijin and Guo, Di and Qu, Xiaobo and Zhou, Jianjun},
  title     = {Fast and Ultra-High Shot Diffusion {MRI} Image Reconstruction with Self-Adaptive {Hankel} Subspace},
  journal   = {Medical Image Analysis},
  year      = {2025},
  volume    = {102},
  pages     = {103546},
  doi       = {10.1016/j.media.2025.103546},
  publisher = {Elsevier}
}

@article{aggarwal2023ensure,
  author    = {Aggarwal, Hemant Kumar and Pramanik, Aniket and John, Maneesh and Jacob, Mathews},
  title     = {{ENSURE}: A General Approach for Unsupervised Training of Deep Image Reconstruction Algorithms},
  journal   = {IEEE Transactions on Medical Imaging},
  year      = {2023},
  volume    = {42},
  number    = {4},
  pages     = {1133--1144},
  doi       = {10.1109/TMI.2022.3224359},
  publisher = {IEEE}
}

@article{xu2025ssflrecon,
  author    = {Xu, Siying and Fr{"u}h, Marcel and Hammernik, Kerstin and Lingg, Andreas and K{"u}bler, Jens and Krumm, Patrick and Rueckert, Daniel and Gatidis, Sergios and K{"u}stner, Thomas},
  title     = {Self-Supervised Feature Learning for Cardiac Cine {MR} Image Reconstruction},
  journal   = {IEEE Transactions on Medical Imaging},
  year      = {2025},
  volume    = {44},
  number    = {9},
  pages     = {3858--3869},
  doi       = {10.1109/TMI.2025.3570226},
  publisher = {IEEE}
}

@article{zhang2022sehlr,
  author    = {Zhang, Xinlin and Lu, Hengfa and Guo, Di and Lai, Zongying and Ye, Huihui and Peng, Xi and Zhao, Bo and Qu, Xiaobo},
  title     = {Accelerated {MRI} Reconstruction With Separable and Enhanced Low-Rank {Hankel} Regularization},
  journal   = {IEEE Transactions on Medical Imaging},
  year      = {2022},
  volume    = {41},
  number    = {9},
  pages     = {2486--2498},
  doi       = {10.1109/TMI.2022.3164472},
  publisher = {IEEE}
}

\end{document}